# A Vision Transformer-Based Approach to Bearing Fault Classification via Vibration Signals

Abid Hasan Zim*, Aeyan Ashraf†, Aquib Iqbal‡, Asad Malik§ and Minoru Kuribayashi¶
*Department of Mechanical Engineering, Aligarh Muslim University, Aligarh 202002, India
†‡Department of Computer Engineering, Aligarh Muslim University, Aligarh 202002, India
§Department of Computer Science, Aligarh Muslim University, Aligarh 202002, India
¶Graduate School of Natural Science and Technology, Okayama University, Okayama 700-8530, Japan
Email: *abid@zhcet.ac.in, {†gj3323,‡gl4127}@myamu.ac.in, §amalik.cs@amu.ac.in, ¶kminoru@okayama-u.ac.jp

*Abstract*—Rolling bearings are the most crucial components of rotating machinery. Identifying defective bearings in a timely manner may prevent the malfunction of an entire machinery system. The mechanical condition monitoring field has entered the big data phase as a result of the fast advancement of machine parts. When working with large amounts of data, the manual feature extraction approach has the drawback of being inefficient and inaccurate. Data-driven methods like Deep Learning have been successfully used in recent years for mechanical intelligent fault detection. Convolutional neural networks (CNNs) were mostly used in earlier research to detect and identify bearing faults. The CNN model, however, suffers from the drawback of having trouble managing fault-time information, which results in a lack of classification results. In this study, bearing defects have been classified using a state-of-the-art Vision Transformer (ViT). Bearing defects were classified using Case Western Reserve University (CWRU) bearing failure laboratory experimental data. The research took into account 13 distinct kinds of defects under 0-load situations in addition to normal bearing conditions. Using the Short Time Fourier Transform (STFT), the vibration signals were converted into 2D time-frequency images. The 2D time-frequency images are then used as input parameters for the ViT. The model achieved an overall accuracy of 98.8%.

*Index Terms*—Bearing-Fault Classification, Vision Transformer (ViT), Deep Learning, Computer vision, Smart manufacturing.

## I. INTRODUCTION

During the last several generations, digitalization has been a major development in production, and it has been assisted by technology related to Artificial Intelligence (AI). This movement has accelerated the advancement of smart manufacturing. Conventional artificial intelligence has been augmented with features specific to current industries, resulting in the development which is referred to as Industrial Artificial Intelligence (IAI), which has emerged as the technological foundation of smart production. Production that is driven by artificial intelligence brings about extraordinary advancements in many parts of closed-loop production systems, including production processes and the logistics of manufactured products [1]. In this quickly evolving technological environment, enterprises throughout the world have expanded the number of methods on their factory floors in order to collect data that might provide them with important insights into their operations [2]. This data-driven strategy helps to acquire knowledge about the machine, and its effective AI analysis may significantly improve operational efficiencies, preventive maintenance, and quality control [3]. Organizations all around the globe are turning to smarter technology as they see the clear advantages of low-cost solutions like AI.

Bearings are vital mechanical components that have several uses in contemporary equipment and the automobile industry. Bearings are often employed in the production of machines and automobiles, primarily to support machine shafts and facilitate rotational motion. Mechanical bearings are crucial components of gear that are prone to wear for a variety of reasons, rendering the equipment unreliable. Consequently, diagnosing issues in bearings is essential for maintaining the safe operation of the machine, which may avoid the emergence of faults and the issue's rapid deterioration, thereby minimizing the decline in profits and ensuring property stability [4], [5].

Bearing health monitoring utilizing data-driven technologies such as machine learning (ML) and deep learning (DL) to anticipate and identify defects in bearings has been the subject of much research. Because faulty bearings may bring down the whole system or cause harm to the entire machine [6]–[8]. In a study Artificial Neural Network (ANN) was used to classify induction motor bearing faults. For categorization, time-domain signal characteristics were extracted [9]. In another study, frequency-domain features serve as the data source for Support Vector Machine (SVM) to identify machinery defects [10]. Traditional ML approaches are mostly built on shallow models, making it difficult to extract non-linear and non-stationary characteristics from the vibration signal [11]. Compared to the conventional machine learning approach, the deep learning method has shown better performance. Researchers used Convolutional Neural Network (CNN), which use vibration signals as their only input data, to diagnose bearing faults [12]. Researchers in one study employed a 1D CNN directly as the input for the raw temporal signal to realizing end-to-end bearing defect detection and achieved great diagnostic accuracy despite noise and various operating loads [13]. Coupling multiscale learning with deep learning methods, different research presented multiscale CNNs for defect detection of wind turbine gearboxes, achieving excellent accuracy. Through the process of transforming signals into two-dimensional (2D) images [14]. DL models with CNN-like architectures, which have dominated computer vision tasks up to this point, have been regarded as the benchmark in image classification and object identification [15]–[17]. CNNs applies convolutional filters to an image in order to extract essential features for the purpose of understanding the object of interest in an image. This is accomplished with the assistance of convolutional operations that cover key properties, including local connection, parameter (weight) sharing, and translation equivarance [18], [19].Unfortunately, CNNs, on the other hand, have poor performance when it comes to learning long-range information, which limits their capability for visual tasks. This is because CNNs have localized receptive fields [20]. On the other hand, the newly suggested Vision Transformer (ViT) promises to be a significant step forward in the deployment of transformer attention models for computer vision tasks, with remarkable performance improvements when compared to the existing models [21]. The ViT uses a transformer architecture on image patches and displays extremely competitive results in image classification. In this paper we have employed Vision Transformer (ViT) on Case Western Reserve University (CWRU) dataset to classify bearing faults. The rest of the paper follows in Section II where the methodology is presented. The results obtained from the model are thoroughly analyzed in Section III. Finally, Section IV includes the conclusion.

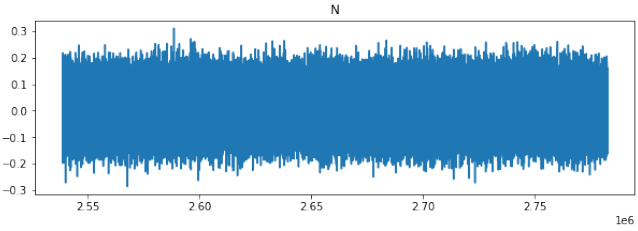

Fig. 1: Raw vibration signal

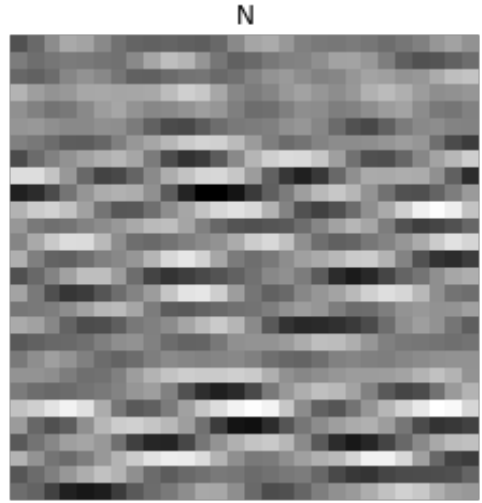

Fig. 2: 2D time-frequency images

## II. METHODOLOGY

### A. Data

Case Western Reserve University (CWRU) bearing failure laboratory experimental data has been utilized in this work to categorize bearing faults using ViT. Electro-discharge machining is used to correct a single point of fault on the bearing. Vibration signals are collected using acceleration sensors. Drive end fault data is sampled at 12000 samples/second and 48000 samples/second, whereas fan end fault data and normal baseline data are also sampled at 12000 samples/second. Consequently, there are four types of data in this public data collection. Each data type is mostly made up of normal, inner race fault, outer race fault, and ball fault. Under loads of 0, 1, 2, and 3 hp, the fault diameters are 0.007 inches, 0.014 inches, and 0.021 inches, respectively [22], [23]. In this work, 13 distinct fault types under 0-load scenario are examined. Along with Normal (N) conditions, fault diameters of 0.007 inches, 0.014 inches, and 0.021 inches have been studied. Fault diameter of 0.007 inches contains ball fault (7_BA), inner race fault (7_IR), outer race fault 1 (7_OR1), outer race fault 2 (7_OR2), outer race fault 3 (7_OR3). Fault diameter of 0.014 inches covers ball fault (14_BA), inner race fault (14_IR), outer race fault 1 (14_OR1). Fault diameter of 0.021 inches holds ball fault (21_BA), inner race fault (21_IR), outer race fault 1 (21_OR1), outer race fault 2(21_OR2), outer race fault 3 (21_OR3). The dataset provided is a raw vibration one-dimensional signal data. To use the ViT with a one-dimensional signal, the data must be pre-processed and the signal must be transformed into an $[m, n, k]$ matrix. The transformation of a signal into a time-frequency map is thus an alternate method. The generic approach for time-frequency analysis comprises Short Time Fourier Transform (STFT), wavelet transformation, etc. STFT is utilized in this study to convert the raw vibration signal (Figure 1) into 2D time-frequency images. Then The 2D time-frequency images (Figure 2) were used for the classification task [24].

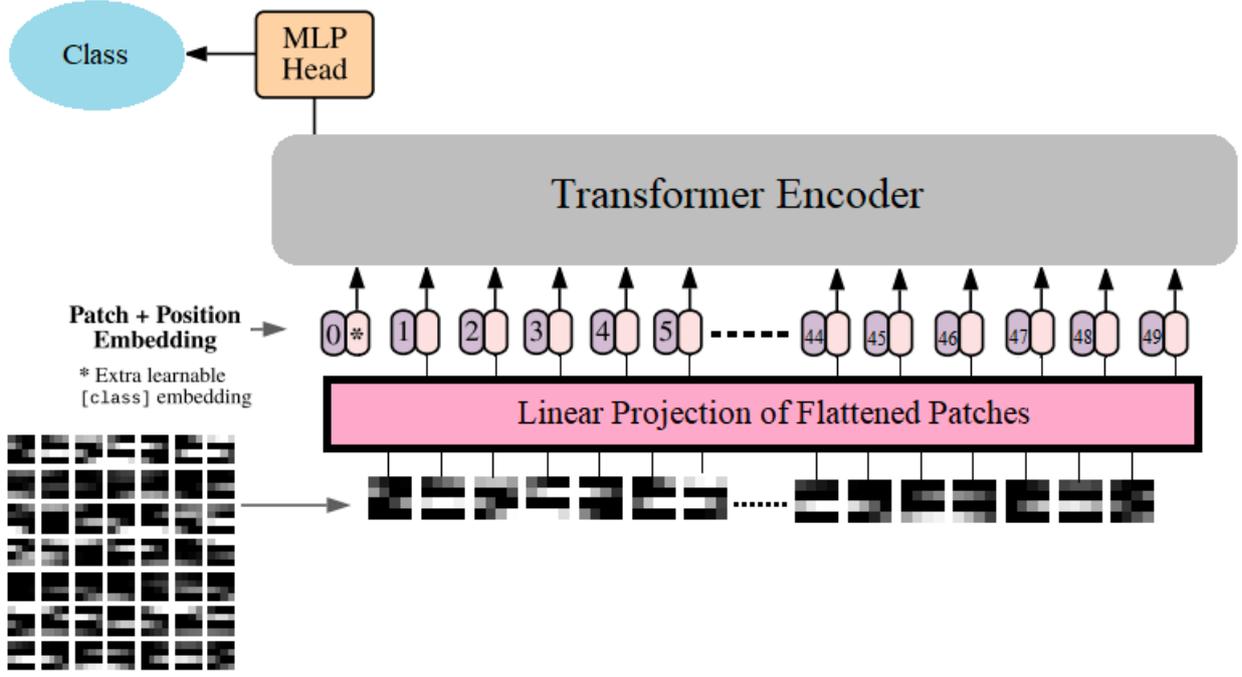

Fig. 3: Model Overview: The input images are split into 49 patches, each patch is linearly embedded, positional embeddings are added and fed into the Transformer Encoder for classification.

*B. Vision Transformer*

The revolutionary transformer architecture has resulted in a significant advancement in the capability for sequence-to-sequence modeling in NLP applications [25], [26]. The phenomenal performance of transformers in NLP has attracted the curiosity of the vision community in determining if transformers can compete effectively against the leading CNNs based architectures in vision applications such as ResNet and EfficientNet [17], [27]. Until relatively recently, the majority of past studies tried to explore transformers centered on merging CNNs with self-attention for computer vision tasks [28]–[31]. While the efficiency of these hybrid techniques is impressive, their computational scalability is inferior when compared to attention-based transformers. The Vision Transformer (ViT), which employs a series of embedded image patches as input to a conventional transformer, represents the first kind of convolution-free transformer that outperforms CNN architectures [21].

The Vision Transformer (ViT) is a pure transformer directly applied to image patch sequences for image classification tasks. It follows the original design of the transformer as closely as possible. To process 2D images, $x \in R^{H \times W \times C}$ is restructured into a series of flattened 2D patches $x_P \in R^{N \times (P^2 \cdot C)}$, where $C$ is the number of channels. The resolution of the main image is $(H, W)$, whereas the resolution of each image patch is $(P, P)$. Therefore, $N = HW/P^2$ is the actual sequence length for the transformer. The outcome of a trainable linear projection (Equation 1), known as patch embeddings, links each vectorized path to the model dimension D since the transformer utilizes constant widths in all of its layers.

$$z_0 = [\text{x}_{\text{class}} ; \text{x}_p^1\mathbf{E} ; \text{x}_p^2\mathbf{E} ; \ldots ; \text{x}_p^N\mathbf{E}] + \mathbf{E}_{pos}, \quad (1)$$

where $\mathbf{E} \in \mathbb{R}^{(P^2 \cdot C) \times D}$ and $\mathbf{E}_{\text{pos}} \in \mathbb{R}^{(N+1) \times D}$.

Researchers add a learnable embedding, similar to BERT's $[class]$ token, to the sequence of embedded patches ($z_o^o = x_{class}$), whose state at the Transformer encoder's output ($z_L^o$) serves as the image representation y (Equation 4). A classification head is connected to $z_L^o$ during both pre-training and fine-tuning. A MLP with a single hidden layer is used to enforce the classification head during pre-training, and a single linear layer is used during fine-tuning.

$$z'_\ell = \text{MSA}\left(\text{LN}\left(z_{\ell-1}\right)\right) + z_{\ell-1}, \quad \ell = 1 \ldots L \quad (2)$$

$$z_\ell = \text{MLP}\left(\text{LN}\left(z'_\ell\right)\right) + z'_\ell, \quad \ell = 1 \ldots L \quad (3)$$

$$y = \text{LN}\left(z_L^0\right) \quad (4)$$

To preserve positional information, position embeddings are incorporated to patch embeddings. Since utilizing more sophisticated 2D-aware position embeddings has not yielded meaningful performance benefits, the researcher employed conventional learnable 1D position embeddings. The encoder receives the generated sequence of embedding vectors as input [21]. The Transformer encoder is made up of layers of multiheaded self-attention (MSA) and MLP blocks (Equations 2,3) [25]. Layernorm (LN) is used before each block, while residual connections are used after each block [32], [33].

To classify the 2D time-frequency images effectively, we propose a transformer-based approach to image classification. The input images are split in 49 patches as compared to the traditional 9 patches approach. Fig. 3 shows the overview of the model

## III. EXPERIMENTS AND RESULTS

### A. Performance Evaluation

In this study the performance of the ViT were evaluated using the accuracy metric. In multi-class classification, accuracy is one of the most prominent measures [34]. The accuracy is calculated as follows:

$$Accuracy = \frac{TP + TN}{TP + FP + TF + FN} \times 100\% \quad (5)$$

where TP: True Positive; TN: True Negative; FN: False Negative, and FF: False Positive.

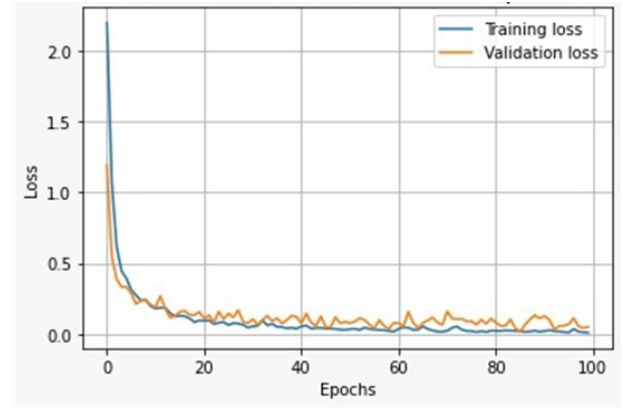

(a)

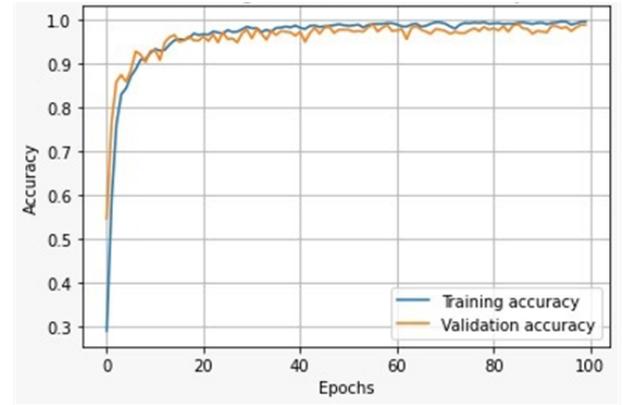

(b)

Fig. 4: (a)Training and Validation losses over epochs (b) Training and Validation accuracy.

### B. Analysis

In this study ViT has been used to classify 13 different types of faults along with normal types in bearing. The overall accuracy obtained from the model is 98.8%. In Fig. 4(a) the Train and validation loss graph depicts the convergence of loss towards zero over the period of 100 epochs and the Fig. 4(b) shows the train and validation accuracy which converges towards one and show the accuracy over the period of 100 epochs. The graph also depicts that the model is fitting optimally on the dataset.

Confusion matrix (Fig. 5) of the model shows overall performance of the model and how our model is predicting on test dataset and also how correctly it is able to classify the image of the same class. It reflects the number of actual and expected values. We can observe that the model was able to classify 14OR1, 21IR, 21OR1, 21OR3, 7BA and 7OR2 faults with 100% accuracy. Overall, the model performs very well.

To show the efficacy of the classification approach for the model suggested in this research, we compared it to classic ML and DL approaches such as Deep Belief Network (DBN) [35], 1-D (time domain) Convolutional Neural Network [36], Stacked Denoising Autoencoder (SDA) [37], Convolutional Neural Network [38]. Table I shows without a doubt that the results of this research's suggested method perform better than those of the methods proposed earlier.

TABLE I: Proposed Scheme's comparison with existing methods.

| Methods | Accuracy |
| --- | --- |
| Deep Belief Network | 89.20% |
| 1-D (time domain) Convolutional Neural Network | 93.30% |
| Stacked Denoising Autoencoder | 95.58% |
| Convolutional Neural Network | 96.48% |
| **Proposed (Vision Transformer)** | **98.80%** |

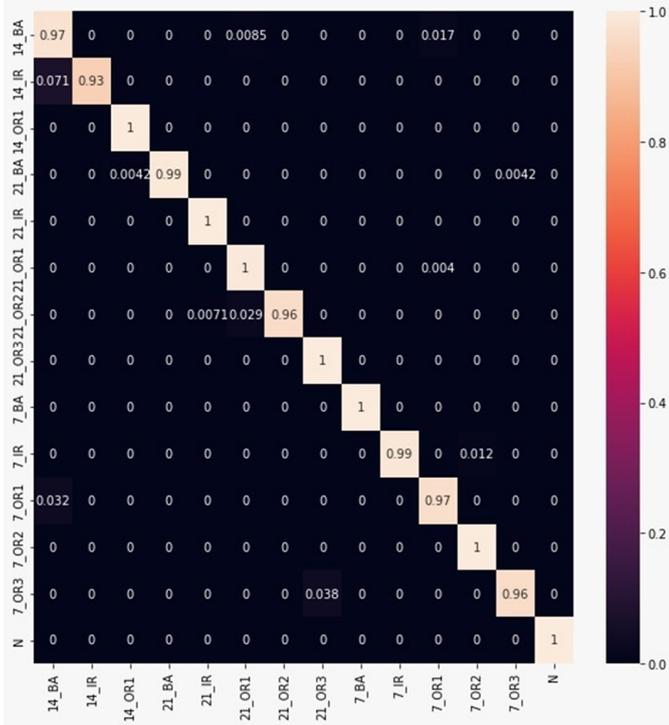

Fig. 5: Confusion matrix of the model.

## IV. CONCLUSION

In this study, state-of-the-art Vision Transformer (ViT) is utilized to classify bearing faults. The ViT uses multi-head self-attention without requiring image-specific biases. The model splits the images into a series of positional embedding patches, which are processed by the transformer encoder. Case Western Reserve University (CWRU) bearing failure laboratory experimental data was used to classify bearing faults. Along with normal bearing conditions, 13 different types of faults under 0-load scenario have been considered in the study. The Short Time Fourier Transform (STFT) was used to transform the raw vibration data into 2D time-frequency images. The 2D time-frequency images were then used as input for classification. The overall accuracy obtained from the model is 98.8%. Among the 13 different fault types, 7 types of faults are correctly classified with 100% accuracy. Overall, the model performs very well.

In the future, more empirical studies can be conducted to better comprehend the ViT's limits, especially with respect to certain faults that are more complicated, such as rolling bearings, gears, and even composite faults. Furthermore, a fixed network design was adopted in this investigation. But the best way to choose parameters is still not clear, especially when a deeper architecture is used or a completely new defect is looked into.

ACKNOWLEDGMENT

This research was supported by the JSPS KAKENHI Grant Number 22K19777, JST SICORP Grant Number JPMJSC20C3 and ROIS NII Open Collaborative Research 2022-22S1402, Japan.

We are very grateful to Asia-Pacific Signal and Information Processing Association for considering our research in their highly distinguished conference "APSIPA ASC 2022"